%% file: main.tex
\documentclass[runningheads,a4paper]{llncs}
\usepackage[margin=1in]{geometry}
\usepackage[utf8x]{inputenc}
\usepackage{amssymb}
\usepackage{graphicx}
\usepackage{booktabs}
\usepackage{array}
\usepackage{url}
\usepackage{subfigure}
\urldef{\mailsb}\path|nathansh@icmc.usp.br, liviacucatto@gmail.com|
\urldef{\mailsc}\path|dbrants@gutennews.com.br, sandra@icmc.usp.br|    
\newcommand{\keywords}[1]{\par\addvspace\baselineskip
\noindent\keywordname\enspace\ignorespaces#1}

\begin{document}

\mainmatter  
\title{
Automatic Classification of the Complexity of Nonfiction Texts in Portuguese for Early School Years}
\titlerunning{
Automatic Classification of Text Complexity}
\author{Nathan Hartmann$^{1}$ \and Livia Cucatto$^{2}$ \and Danielle Brants$^{2}$ \and Sandra Aluísio$^{1}$}
\institute{$^{1}$Interinstitutional Center for Computational Linguistics (NILC)\\Institute of Mathematical and Computer Sciences\\University of São Paulo\\
$^{2}$GUTEN EDUCAÇÃO E TECNOLOGIA LTDA.\\
\mailsb\\
\mailsc\\}

\maketitle

\begin{abstract}
Recent research shows that most Brazilian students have serious problems regarding their reading skills. The full development of this skill is key for the academic and professional future of every citizen. Tools for classifying the complexity of reading materials for children aim to improve the quality of the model of teaching reading and text comprehension. For English, Feng’s work \cite{feng2010} is considered the state-of-art in grade level prediction and achieved 74\% of accuracy in automatically classifying 4 levels of textual complexity for close school grades. There are no classifiers for nonfiction texts for close grades in Portuguese. In this article, we propose a scheme for manual annotation of texts in 5 grade levels, which will be used for customized reading to avoid the lack of interest by students who are more advanced in reading and the blocking of those that still need to make further progress. We obtained 52\% of accuracy in classifying texts into 5 levels and 74\% in 3 levels. The results prove to be promising when compared to the state-of-art work. 

\end{abstract}

\input{introduction}

\input{related_works}

\input{manual_annotation}

\input{experiments}

\input{conclusions}

\bibliographystyle{splncs03}
\bibliography{referencias}

\end{document}

%% file: introduction.tex
\section{Introduction}

According to data collected by the Organisation for Cooperation and Economic Development (OECD) in the Programme for International Student Assessment (PISA)\footnote{Available at \url{oecd.org/education/PISA-2012-results-brazil.pdf}}, Brazilian students have serious problems regarding their reading skills. The most recent survey, carried out in 2012, showed results for Brazil below the average of the countries surveyed. 49.5\% of Brazilian students did not reach the levels considered minimum in reading, which means that, at best, they can only recognize themes of simple and familiar texts. Furthermore, only 0.5\% of Brazilian students reached maximum reading levels, which means that only one in every 200 young people in Brazil is able to deal with complex texts and perform in-depth analysis on such texts. More negative numbers were seen in the Brazilian National High School Exam (ENEM – \textit{Exame Nacional do Ensino Médio}) in 2014: from the 6.1 million students who did the exam, 529 flunked the composition. Experts stated that most students do not even understand the wording of the question. Only 250 students, equivalent to 0.004\%, aced the composition.

The development of reading skills has long been related to success in future academic and professional activities.  Aimed at raising the quality of the teaching model for reading and text comprehension in this country and trying to close some gaps in Brazilian public policies for education, many features and computer systems for the Brazilian Portuguese have been launched recently.
An example is the First Book Project (\textit{Projeto Primeiro Livro})\footnote{Available at \url{primeiro-livro.com}}, which helps children and young people from public schools to learn grammar, spelling and develop narratives. Another example is the Victor Civita Foundation, sponsored by the publishing house Abril, which supports teachers, school managers and public policy makers of Elementary Education with lesson plan search engines, social network for educators to exchange experience and share knowledge, and a resource bank for classes\footnote{Available at \url{rede.novaescolaclube.org.br}}.

Currently, in Brazil, the elementary school is divided into two stages - 1st to 5th year, and 6th to 9th year. The National Curriculum Parameters (1998), however, divide these two stages into four cycles. In this article, we focus on the end of the first cycle - 3rd year -, and the second and third cycles - 4th/5th and 6th/7th years because they are fundamental for students to achieve adult reading comprehension. 

There are some tools for Brazilian Portuguese such as the Flesch Index \cite{martins1996}, which is adapted for Portuguese and used in the Microsoft Word, and mainly the Coh-Metrix-Port and AIC, developed in the PorSimples project \cite{aluisio2010}, whose goal is to simplify Web texts for people with poor literacy levels. These tools, however, do not meet the needs of educators in the classroom: there are no classifiers able to discriminate the level of complexity of each year focus of this study – 3rd to 7th years, using metrics of the many language levels.

For the English language, there are tools for classifying reading materials for children used in US schools, based on both quantitative data such as Lexile\footnote{Available at \url{lexile.com}} \cite{lennon2004} \cite{stenner1996} and better informed such as Text Easability Assessor (TEA)\footnote{Available at \url{tea.cohmetrix.com}} that uses Coh-Metrix \cite{graesser2011} \cite{graesser2004} metrics. 

In this article, we present the process of features development and training of a classifier based on machine learning to automatically distinguish five levels of textual complexity to support the selection of texts for students of a given class. Here, we use grade levels, which indicate the number of years of education required to completely understand a text, as a proxy for reading difficulty, the same way as \cite{feng2010}. However, we understand that there can be a great diversity of competences, abilities and background knowledge regarding reading in a same classroom.

In Section \ref{sec:related_works} we present some recent work on automatic readability assessment of grade levels. In Section \ref{sec:manual_annotation} we present the manual annotation criteria and the process of manual annotation of our corpus. In Section \ref{sec:experiments} we present the experiments carried out and the results obtained on 5 grade levels and on combining adjacent levels, achieving best results on 3 classes. Finally, in Section \ref{sec:conclusions} we present our final remarks and future work. 

%% file: related_works.tex
\section{Related Work}
\label{sec:related_works}

In recent years, the interest in building automatic classifiers of text complexity has increased. Although the English language is a highlight in this topic \cite{collins2011}  \cite{graesser2011} \cite{lopucki2014} \cite{sheehan2013}, it has served as base for other languages to develop their own classifiers, such the French \cite{franccois2014}, Italian \cite{dell2014}, Spanish \cite{san2014}, German \cite{hancke2012}  \cite{vajjala2014}, Arabic \cite{forsyth2014} and Portuguese \cite{aluisioetal2010} \cite{curto2014}. Automatic classifiers of text complexity have various applications, as follows: teaching a second language \cite{curto2014}, reading and comprehension for poor literacy readers \cite{aluisio2010}, legal and scientific texts and as a first step in building Text Simplification Systems \cite{aluisioetal2010}.

Readability studies are an area of great interest for language teaching, particularly in building materials for reading and learning vocabulary. The studies in this area allow to establish a scale of difficulty levels of texts used to assess students. Generally, in elementary levels of education, teachers acknowledge that giving reading materials not suitable for the students’ level impairs their learning, discouraging them \cite{fulcher1997}. 

Curto \cite{curto2014} developed a system to extract linguistic features and a text classifier to teach Portuguese as a second language. The motivation presented by the author is the need of selecting texts for language teaching, which is done manually. 

The Coh-Metrix-Port 2.0\footnote{Available at \url{nilc.icmc.usp.br/coh-metrix-port}}, an adaptation of the Coh-Metrix developed in the PorSimples project \cite{aluisioetal2010}, currently provides 48 metrics that enable the analysis of lexical, morphosyntactic, syntactic (chunking), semantic and discursive features \cite{scartonaluisio2010}. The AIC tool, with 39 metrics \cite{maziero2008aic}, covers the lack of syntactic analysis (full parsing) in the Coh-Metrix-Port. Scarton and Aluísio \cite{scartonaluisio2010} evaluated the first version of the Coh-Metrix-Port tool (with 38 metrics) comparing written texts for adults with written texts for children, considering only two levels: simple texts and complex ones related to the journalistic and scientific dissemination genre. It is worth noting that a simple measure such as the Flesch Index and its components results in a SVM classifier with polynomial kernel with 82.5\% accuracy, while the Coh-Metrix-Port increased accuracy to 92\% and the measures altogether resulted in 93\% of accuracy. 

The work most related to ours is for the English language \cite{feng2010} and classifies textual complexity using a corpus of magazines for elementary and high school students (Weekly Reader Corpus\footnote{Available at \url{www.weeklyreader.com}} that has texts for elementary school students labeled with grade levels, which range from 2 to 5). Their best results were obtained by group-wise add-one-best feature selection, resulting in 74\% classification accuracy, with 273 features selected, including language modeling features, syntactic features, PoS features, traditional readability metrics, and out-of-vocabulary features.

%% file: manual_annotation.tex
\section{Corpus and Manual Annotation on Grade Levels}
\label{sec:manual_annotation}

\subsection{Description of grade levels and the problem } 

In recent years, the Brazilian government has been working on a systematization of the education policy in an attempt to unify the curricula methods and content for schools and teachers all over Brazil to speak the same language. The \textit{Provinha Brasil}\footnote{\textit{Provinha Brasil} is a test to evaluate how much children have learned about Portuguese and Mathematics subjects. Available at \url{provinhabrasil.inep.gov.br}}, the state assessment tests (e.g., SARESP\footnote{Available at \url{http://www.educacao.sp.gov.br/saresp}} in the state of São Paulo) and even the ENEM (National High School Exam) are attempts to direct education professionals to the same educational setting. However, it is still not clear for teachers, especially for elementary school ones, how to distribute such content by school year, especially when it comes to reading. In addition, in Brazil, there is an extremely diverse learning scenario in the same grade. The insertion of dictionaries in grade levels by the National Textbook Program (PNLD) \cite{krieger2012} since 2006 shows a change, albeit slow, in the Brazilian educational system. 

Building a five-level classifier is in line with this emerging educational scenario. For the 3rd, 4th and 5th years (\textit{Ensino Fundamental I}) and the 6th and 7th years of the elementary school (\textit{Ensino Fundamental II}), we can measure the complexity of texts and, thus, meet the diversity in reading comprehension. 

The creation basis was: the National Curriculum Parameters (PCNs) (1998), the descriptors of Prova Brasil\footnote{\textit{Prova Brasil} is a test to evaluate the quality of the educational brazilian system. Available at \url{http://portal.mec.gov.br/prova-brasil}}, analysis of textbooks, articles in
the psycholinguistics area \cite{cimadon2012} \cite{flor2014} \cite{giangiacomo2008} \cite{maia2001} \cite{maia2011} \cite{maia2005} \cite{navas2009} \cite{oreilly2004} \cite{desalles2006} and language acquisition \cite{kato1985} \cite{kato1986}, and the knowledge of linguists with experience in Education and the Portuguese language (phonology, morphology, syntax, semantics and discourse). 

With respect to PCNs, one way to measure these skills was to create descriptors that synthesized the competencies and skills. Such descriptors are used as reference matrix for Prova Brasil. The Portuguese language test assesses only reading skills, represented by 21 descriptors for the 9th year and by 15 descriptors for the 5th year, divided into six groups: (1) Reading procedures; (2) implications of support, gender and/or enunciator in the text comprehension; (3) Relationship between texts; (4) Coherence and cohesion in text processing; (5) Relations between expressive features and effects of meaning; and (6) Linguistic variation. However, neither the PCNs nor the descriptors distinguish five levels. On the other hand, it is known that each grade level has a specific curriculum and, therefore, its difficulties and expected progress. One way to obtain a more objective division by grade levels was to resort to textbooks. All of them indicate the content to be taught and bring nonfiction texts.

\subsection{Corpus and selection of texts for annotation}
\label{sec:corpuseselecao}

In order to build the corpus, we search for pre-selected texts in terms of complexity levels, using the following sources: SARESP and textbooks. We obtained only 72 texts, distributed in five levels, from SARESP tests, given limitations such as they do not cover all school years; they are generally applied once a year; the test contains several textual genres – that is, there are few informative texts; and, above all, not all texts are available online. Considering the difficulties above and knowing the importance of a large amount of data to machine learning techniques, we turned to textbooks as our main source of texts. Experts selected 178 informative texts from Portuguese language textbooks. Therefore, we equally distributed 50 texts in each level, totaling 250.

Because of the small amount of texts which had some level information, new sources, not previously classified, were included in the corpus: NILC corpus\footnote{Available at \url{nilc.icmc.usp.br/nilc/images/download/corpusNilc.zip}}, \textit{Ciência Hoje das Crianças} (CHC)\footnote{Available at \url{chc.cienciahoje.uol.com.br}}, \textit{Folhinha}\footnote{Available at \url{www.folha.uol.com.br/folhinha}}, \textit{Para Seu Filho Ler}\footnote{Available at \url{zh.clicrbs.com.br/rs}} and \textit{Mundo Estranho}\footnote{Available at \url{mundoestranho.abril.com.br}}, which currently contains 7,645 texts compiled, whose sources distribution is shown in Table \ref{tab:corpusdistribution}. Among the seven sources, the one that presents great diversity of textual type and gender is textbooks, since the purpose of this type of source is to present the student with all existing genres and types – we found from simple expository texts to more complex structures such as argumentative texts very common in the editorial genre; the same textual amplitude is seen in SARESP tests\footnote{Available at \url{sites.google.com/site/provassaresp}}. Although the NILC corpus is also composed of textbooks, its texts generally have three text types: descriptive, narrative and expository. However, CHC, \textit{Folhinha} and \textit{Mundo Estranho} are similar: they present, in most cases, dialogues; varied text types in the same text; and the predominance of a particular type. These different possibilities of textual occurrence increase the challenge of building the curricula (see Section \ref{sec:criteriosanotacao}) and, therefore, the classification system. So far, 1,456 texts have been annotated by a sole linguist.

\begin{table}[!ht]
	\center
	\begin{tabular}{c|c|c|c|c|c|c}
      \toprule
  & NILC & SARESP  & \textit{Ciência Hoje}& \textit{Folhinha}  & \textit{Para seu Filho Ler}  & \textit{Mundo} \\
		  Textbooks & corpus  & tests & \textit{das Crianças}   & Issue of Folha  & Issue  of & \textit{Estranho}\\
       & & & & de São Paulo &  Zero Hora & \\
      \midrule
		492 & 262 & 72 & 2.589 & 308 & 166 & 3.756\\
      \bottomrule
	\end{tabular}

	\caption{Distribution of texts by source.}
\label{tab:corpusdistribution}
\end{table}

\subsection{Annotation criteria}

\label{sec:criteriosanotacao}

The first annotation grid built relied on textbook curricula, which has linguistic phenomena organized by grade levels. From this basis, the contact with texts targeted to school years and the knowledge of linguists, we kept on improving the grid. We should emphasize that although the school introduces linguistic elements in certain years, children can already understand and produce them long before being exposed to them in the educational system. Hence, the need to link different sources of knowledge.

Another challenge lies in the text type diversity found in informative texts, namely: narrative, descriptive, injunctive, expository and argumentative \cite{bakhtin1992}. Such text types have different structures, but they may still be in the same reading comprehension level. Thus, for example, a mostly injunctive text may have the same level of complexity as a text that is mostly descriptive. Structural possibilities were and are still considered in the grid detailing. 

Linguistic and non-linguistic elements are divided into six groups: morphological, lexical, syntactic, textual, punctuation and semantic and reader’s commonsense knowledge. The first one corresponds to linguistic elements in the morphological level such as verb endings, affixes and grammatical categories; the second brings together linguistic phenomena connected to vocabulary and semantic relationships such as synonymy, antonymy, polysemy, among others; the syntactic group highlights the types of clauses present in the texts, how they are organized within the sentence, the paragraph, the order and size of constituents; with regard to text metrics, the main focus is cohesion: the type of cohesion used and the elements used for this end. The Punctuation and Semantic and reader’s commonsense knowledge complement the previous ones: this maps the punctuation richness and the other is an attempt to capture the semantic and world knowledge of the reader, so far, by means of named entities. 

%% file: experiments.tex
\section{Experiments}
\label{sec:experiments}

\subsection{Preliminary Experiments: using language independent features}

The manual annotation process started focusing on a balanced sample of 971 texts in 5 levels of textual complexity, from the 3rd to 7th grade levels, mapped here from level 1 to 5. The distribution of our initial data set is as follows: 208 texts of level 1, 185 texts of level 2, 196 texts of level 3, 191 texts of level 4 and 191 texts of level 5. For this set of texts, we extracted the following 10 features list we call ``simple statistics feature'': Flesch-Kincaid Grade Level index, the average sentences per paragraph, average words per sentence, number of paragraphs, number of sentences, number of words in the text, type-token ratio, number of simple words matching the dictionary of simple words to youngsters \cite{biderman2003}, incidence of punctuation and diversity of punctuation. All of these features are independent of language, except for the dictionary of simple words, but it is easy to find it for many languages. When performing a 10-fold cross-validation experiment on the initial data set, with an SVM classifier\footnote{It was used a libsvm implementation of SVM classifier.} with linear kernel and C=1, we obtained 52\% of accuracy (+/- 14). It is worth noting that the 3 features best classified by the recursive feature elimination (RFE) process for selecting features were the Flesch-Kincaid, the number of paragraphs in the text and the diversity of punctuation.

\subsection{Increasing the Number of Features and Data}

Keeping the size of the initial corpus, we decided to increase our features set to better represent differences among the textual levels. Table \ref{tab:allfeatures} maps the features implemented in 6 linguistic categories used for corpus annotation, described in Section \ref{sec:criteriosanotacao}. 
Table \ref{tab:allfeatures} shows a total of 108 features: (i) 52 Coh-Metrix-Port features 
2.0\footnote{Available at \url{http://143.107.183.175:22680}}, (ii) 32 AIC Features, (iii) two features based on the lists of positive and negative words of the LIWC - Dictionary for Sentiment Analysis\footnote{Available at \url{http://143.107.183.175:21380/portlex/index.php/en/liwc}}, 14 features about Named Entities, calculated on the flat output of the PALAVRAS parser \cite{bick2000}, and (v) 8 new features on Verbs Incidence implemented especially for this work comprising Portuguese verb tenses and moods. Some features were duplicated on Table \ref{tab:allfeatures} because they use information from many linguistic categories.

\begin{table}[!ht]
\center
\scriptsize

\scalebox{.85}{
\begin{tabular}{lll}
\toprule
\textbf{Morphological Features} &\\
\midrule
Inc. of Indicative mood (preterite perfect tense) & Mean syllables per content word & Inc. of Imperative mood\\
Inc. of Indicative mood (imperfect	 tense)  & Inc. of Indicative mood (future tense) & Inc. of Subjunctive mood\\
Inc. of Indicative mood (pluperfect tense) & Inc. of Indicative mood (present tense) & Flesch index \\
Inc. of Indicative mood (future of the past tense) & \\

\end{tabular}
}

\scalebox{.99}{
\begin{tabular}{lll}
\toprule
\textbf{Lexical Features} & & \\
\midrule
Adjective incidence & Adverb incidence & Content word incidence\\
Flesch index & Function word incidence & Mean words per sentence\\
Noun incidence & Number of Words & Verb incidence\\
Content words frequency (BP) & Min among content words freq & Mean hypernyms per verb\\
Brunet Index & Honore Statistic & Mean pronouns per noun phrase\\
Type to token ratio & Ambiguity of adjectives & Ambiguity of adverbs\\
Ambiguity of nouns & Ambiguity of verbs & Words before Main Verb\\
Inc. of Prepositions Per Clauses & Inc. of Prepositions Per Sentence & \\
\end{tabular}
}

\scalebox{0.713}{
\begin{tabular}{lll}
\toprule
\textbf{Syntactic Features} & & \\
\midrule
Mean Clauses per Sentence & Mean pronouns per noun phrase & Modifiers per Noun Phrase\\
Noun Phrase Inc. & Mean Adverbial Adjunct Per Phrase & Inc. of Coordinate Clauses\\
Mean Apposition Per Clause & Inc. of Gerund Verbs & Inc. of Infinitive Verbs\\
Inc. of Verbals & Inc. of Coordinate Clauses & Mean of Clauses Per Sentence\\
Inc. of Initiating Subordinate Clauses & Inc. of Participle Verbs & Inc. of Passive Sentences\\
Inc. of Prepositions Per Clauses & Inc. of Prepositions Per Sentence & Inc. of Relative Clauses\\ Inc. of  Sentences With 5 Clauses & Inc. of  Sentences With Four Clauses & Inc. of  Sentences With 1 Clause\\
Inc. of  Sentences With 7 or More Clauses & Inc. of  Sentences with 6 Clauses & Inc. of  Sentences With 3 Clauses\\
Inc. of  Sentences With 2 Clauses & Inc. of  Sentences With Zero Clauses & Inc. of Subordinate Clauses\\
Inc. of Imperative mood & Inc. of Subjunctive mood  & Inc. of Indicative mood (future tense)\\
Inc. of Indicative mood (preterite tense) & Inc. of Indicative mood (pluperfect tense) & Inc. of Indicative mood (present tense)\\
Inc. of Indicative mood (preterite perfect tense)  & Inc. of Indicative mood (future of the past tense) & \\
\end{tabular}
}

\scalebox{.91}{
\begin{tabular}{lll}
\toprule
\textbf{Textual Features} & & \\
\midrule
Inc. of ANDs & Inc. of IFs & Inc. of ORs\\
Inc. of negations & Logic operators Inc. &  Inc. of connectives\\
Inc. of additive negative connec. & Inc. of additive positive connec. & Inc. of causal negative connec.\\
Inc. of causal positive connec. & Inc. of logical negative connec. & Inc. of logical positive connec.\\
Inc. of temporal negative connec. & Inc. of temporal positive connec. & Adjacent anaphoric references\\
Anaphoric references & Adjacent argument overlap & Argument overlap\\
Adjacent stem overlap & Stem overlap & Adjacent content word overlap\\
Inc. of Ambiguous Discourse Markers & Inc. of Discourse Markers & Incidence of Pronouns\\
Inc. of 1st Person Poss. Pronouns & Inc. of 1st Person Pronouns & Inc. of 2nd Person Poss. Pronouns\\
Inc. of 2nd Person Pronouns & Inc. of 3th Person Poss. Pronouns & Inc. of 3th Person Pronouns\\
\end{tabular}
}

\scalebox{1.01}{
\begin{tabular}{lll}
\toprule
\textbf{Punctuation Features} & &\\
\midrule
Punctuation diversity in a text & Number of Paragraphs in a text & Punctuation incidence in a text\\ 
Number of sentences in a text & Flesch index & \\ 
\end{tabular}
}

\scalebox{0.83}{
\begin{tabular}{ll}
\toprule
\textbf{Semantic and reader’s commonsense knowledge} & \\
\midrule
Inc. of LIWC Negative Words & Inc. of LIWC Positive Words\\ Inc. of Concrete Moving Entities in Sentences & Inc. of Concrete Moving Entities in Text\\ Inc. of Concrete Non-Moving Entities in Sentences & Inc. of Concrete Non-Moving Entities in Text\\
Inc. of Human Named Entities in Sentences & Inc. of Human Named Entity Sentence\\
Inc. of Named Entities in Sentences & Inc. of Named Entities in Text\\
Inc. of Non-Human Anim. Moving Entities in Sentences & Inc. of Non-Human Anim. Moving Entities in Text\\
Inc. of Non-Human Anim. Non-Moving Entities in Sentences &
Inc. of Non-Human Anim. Non-Moving Entities in Text\\ Inc. of Topological Entities in Sentences & Inc. of Topological Entities in Text\\
\bottomrule
\end{tabular}
}

\caption{Full set of 108 features currently been used.}
\label{tab:allfeatures}
\end{table}

By repeating the experiment with the same fold and SVM settings for the new set of 108 features, we obtained 56\% of accuracy (+/-13). We know it is difficult to have statistical learning in a small dataset such as the initial dataset. Therefore, we use the Active Learning Approach \cite{tong2002} for selecting new instances for annotation, so that the new instances are those that are most difficult for our classifier to label. Thus, we use the distance of texts from SVM separating hyperplanes as criteria for selecting instances for annotation. The closer an instance is from the separating hyperplanes, there is greater indecision in classifying that instance. Therefore, when we label this text manually, we believe we are helping the classifier to better define the existing limits between classes. 

We performed four steps to select texts for annotation, where each step selected the 100 most complex texts for SVM. The texts that could not be processed due to parsing problems were removed. The results are shown in Table \ref{tab:active_learning}. They show that even when we select the texts in which the classifier has greater indecision in classifying, the SVM has not yet been able to define a boundary between the classes, which led to lower accuracy in classifying data. This shows that there is a mix between classes so that the 108 current features are not able to correctly distinguish the five levels manually annotated. 
Finally, we conducted a stage of selecting the 100 most easily annotated texts (those with greater distance from SVM separating hyperplanes) in order to contrast with the current distribution of data and the accuracy obtained. We obtained a set of 1,456 texts with the following distribution: 242 texts of level 1, 313 texts of level 2, 338 texts of level 3,287 texts of level 4 and 276 texts of level 5. The accuracy obtained when performing a 10-fold cross-validation experiment with linear kernel SVM and C=1 was 52\% (+/- 15). 

\begin{table}[!ht]
	\center
	\begin{tabular}{lcc}
    \toprule
	\textbf{Step} & \textbf{Texts} & \textbf{Accuracy}\\
    \midrule
	First & 1,070 & 53 (+/-11)\\
    Second & 1,169 & 50 (+/-14)\\
    Third & 1,268 & 51 (+/-13)\\
    Forth & 1,364 & 50 (+/-15)\\
    \bottomrule
\end{tabular}
	\caption{Selection of texts via Active Learning and accuracy obtained from SVM}
\label{tab:active_learning}
\end{table}

\begin{figure}[!ht]
 \centering
\includegraphics[scale=.7, trim= 0 1.5cm 0 0]{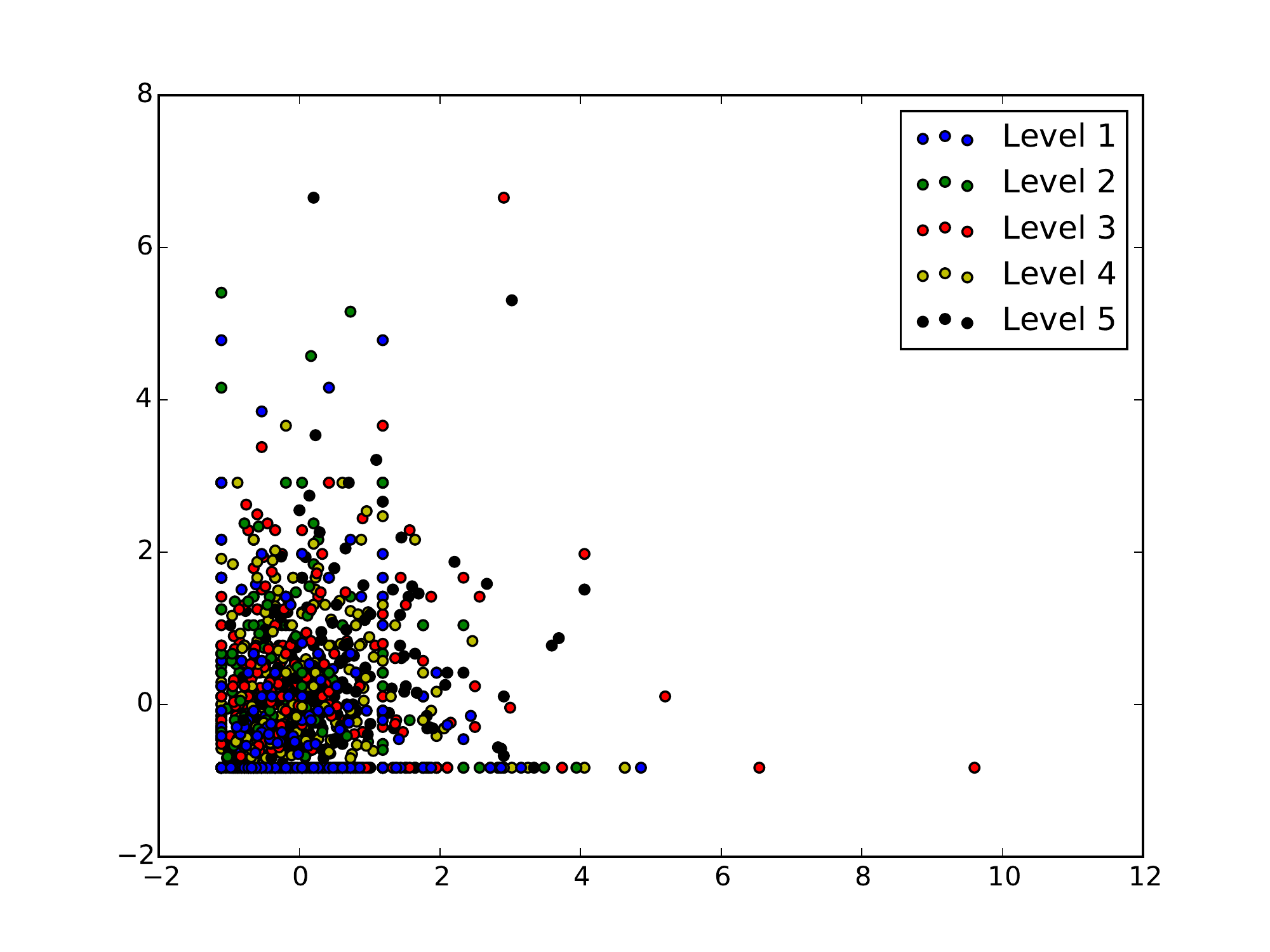}
 \caption{$\mathbb{R}$$^{2}$ distribution of our 1,456 texts with the 2 most significant features. X-axis represents Incidence of Indicative mood (Preterit perfect tense) and Y-axis Incidence of additive negative connectives. Data scaling with mean 0 and standard deviation 1.}
\label{fig:data_distribution}
\end{figure}

This slight improvement in performance shows us that, in fact, there is a set of complex texts that the classifier cannot handle: due to either lack of discriminative features or lack of data for training (see confusion matrix on Table \ref{tab:confusion_matrix}). The problem can also consist in human annotation errors. To evaluate that we performed a double-blind annotation of a random sampling of 100 texts. We obtained a Kappa score of 0.528 that represents a moderate agreement on Landis and Koch scale \cite{landiskoch1977}. This agreement suggests that the manual annotation process and the labeled data should be reviewed because, as Hovy and Lavid says, ``if humans can agree on something at N\%, systems will achieve (N-10)\%'' \cite{hovy2010}. In addition to the confusion matrix, we can see in Figure \ref{fig:data_distribution} the axes that represent the two most discriminative features of the 44 selected by the RFE method of feature selection, and that there is, in fact, a mixture in the features space, particularly between the 2-3, 3-4-5, and 4-5 levels. This scenario will be hardly separated by SVM.

\begin{table}[!ht]
	\center
	\begin{tabular}{rrrrrr}
    \toprule
    & \textbf{Level 1} & \textbf{Level 2} & \textbf{Level 3} & \textbf{Level 4} & \textbf{Level 5}\\
    \midrule
    \textbf{Level 1} & 182 & 45 & 9 & 4 & 2\\
    \textbf{Level 2} & 36 & 160 & 102 & 14 & 1\\
    \textbf{Level 3} & 11 & 99 & 170 & 39 & 19\\
    \textbf{Level 4} & 6 & 13 & 79 & 118 & 71\\
    \textbf{Level 5} & 3 & 5 & 28 & 60 & 180\\
    \bottomrule
\end{tabular}
	\caption{Confusion matrix of a 10-fold cross-validation experiment on our dataset.}
\label{tab:confusion_matrix}
\end{table}

Feng’s work \cite{feng2010} addresses 4 levels of difficulty, reaching the state-of-art 74\% of accuracy in English. Our experiments with fewer classes showed that, when joining classes 2 and 3, we achieved 65\% (+/- 15) of accuracy, and by joining classes 4 and 5, we achieved 63\% (+/- 11) of accuracy. By simultaneously joining class 2 with class 3 and 4 with 5, we reached the 74\% of accuracy achieved by the state of art. This division of grade levels better reflects the division into cycles indicated by the PCNs (1998).

%% file: conclusions.tex
\section{Discussion and Future Work}
\label{sec:conclusions}

Our work presents the first efforts to automatically classify Portuguese texts into 5 close grade levels. The literature shows that this task is complex and, in this sense, our results are promising. We also understand that, despite the number of features used is 40\% of the 273 features used in the state-of-art work for the English language \cite{feng2010}, there is a high rate of mixed data, especially in the central levels 4-6. Our selection of features brought 44 of the 108 features used in this work, obtaining 52\% (+/- 15) of accuracy. This selection brings features to meet 5 out of 6 linguistic groups that model the manual annotation, for example: Flesch Index for the Morphological category; Ambiguity of adjectives and Incidence of Adverbs for the Lexical category; Mean Apposition Per Clause for the Syntactic category; Adjacent content word overlap and Incidence of Negative Additive Connective for the Textual category; Incidence of Human Named Entity in Text for the Semantic and reader’s commonsense knowledge. By reducing the classification to 3 levels of textual complexity, we achieved 74\% of accuracy - as obtained by the state-of-art work for the English language that focuses on 4 levels. 

As future work, we indicate two fronts of efforts:(i) the re-annotation of the corpus by a second annotator, using the manual annotation developed to check discrepancies; (ii) the addition of features in the six categories of linguistic elements that were used for manual classification of texts. We will replicate 6 out-of-vocabulary features described in \cite{feng2010}. For each text in our final corpus, these 6 features are computed using the most common 100, 200 and 500 word tokens and types based on texts from 3th grade. Also, we will implement successful features for the English language, cited by \cite{petersen2009}, such as average sentence length and features from the language model of our corpus. Moreover, and more importantly, we will implement a text type classifier to distinguish the text types occurring in our corpus. As the features of each text in our corpus are being annotated and there is a corpus annotated with text types in the Lácio-Web project \cite{aluisio2004} we will be able to better understand the correlations between text types and the others features for readability assessment in our project.

%% file: main.bbl
\begin{thebibliography}{10}
\providecommand{\url}[1]{\texttt{#1}}
\providecommand{\urlprefix}{URL }

\bibitem{aluisioetal2010}
Aluisio, S., Specia, L., Gasperin, C., Scarton, C.: Readability assessment for
  text simplification. In: Proceedings of the NAACL HLT 2010 Fifth Workshop on
  Innovative Use of NLP for Building Educational Applications. pp. 1--9.
  Association for Computational Linguistics (2010)

\bibitem{aluisio2004}
Alu{\'\i}sio, S.M., Pinheiro, G.M., Manfrin, A.M., de~Oliveira, L.H.,
  Genoves~Jr, L.C., Tagnin, S.E.: The l{\'a}cio-web: Corpora and tools to
  advance brazilian portuguese language investigations and computational
  linguistic tools. In: Proceedings of LREC. pp. 1779--1782 (2004)

\bibitem{aluisio2010}
Alu{\'\i}sio, S.M., Gasperin, C.: Fostering digital inclusion and
  accessibility: the porsimples project for simplification of portuguese texts.
  In: Proceedings of the NAACL HLT 2010 Young Investigators Workshop on
  Computational Approaches to Languages of the Americas. pp. 46--53.
  Association for Computational Linguistics (2010)

\bibitem{bakhtin1992}
Bakhtin, M.: Est{\'e}tica da cria{\c{c}}{\~a}o verbal. Livraria Martins Fontes
  (2003)

\bibitem{bick2000}
Bick, E.: The Parsing System ``Palavras'': Automatic Grammatical Analysis of
  Portuguese in a Constraint Grammar Framework. Aarhus University Press Aarhus
  (2000)

\bibitem{biderman2003}
Biderman, M.T.C.: Dicion{\'a}rios do portugu{\^e}s: da tradi{\c{c}}{\~a}o {\`a}
  contemporaneidade. ALFA: Revista de Lingu{\'\i}stica  47(1) (2003)

\bibitem{cimadon2012}
Cimadon, {\'E}.: Fun{\c{c}}{\~o}es executivas em crian{\c{c}}as com dificuldade
  de leitura  (2012)

\bibitem{collins2011}
Collins-Thompson, K., Bennett, P.N., White, R.W., de~la Chica, S., Sontag, D.:
  Personalizing web search results by reading level. In: Proceedings of the
  20th ACM international conference on Information and knowledge management.
  pp. 403--412. ACM (2011)

\bibitem{curto2014}
Curto, P.: Classificador de textos para o ensino de português como segunda
  língua. Master's thesis, Universidade Técnico Lisboa, Portugal (2014)

\bibitem{dell2014}
Dell’Orletta, F., Venturi, G., Cimino, A., Montemagni, S.: T2k2: System for
  automatically extracting and organizing knowledge from texts. In: Proceedings
  of the 9th International Conference on Language Resources and Evaluation
  (LREC’14) (2014)

\bibitem{feng2010}
Feng, L., Jansche, M., Huenerfauth, M., Elhadad, N.: A comparison of features
  for automatic readability assessment. In: Proceedings of the 23rd
  International Conference on Computational Linguistics: Posters. pp. 276--284.
  COLING '10, Association for Computational Linguistics (2010)

\bibitem{flor2014}
Flor, M., Klebanov, B.B.: Associative lexical cohesion as a factor in text
  complexity. International Journal of Applied Linguistics  165(2),  223--258
  (2014)

\bibitem{forsyth2014}
Forsyth, J.N.: Automatic Readability Detection for Modern Standard Arabic.
  Master's thesis, Brigham Young University, United States

\bibitem{franccois2014}
Fran{\c{c}}ois, T.: An analysis of a french as a foreign language corpus for
  readability assessment. NEALT Proceedings Series Vol. 22 pp. 13--32 (2014)

\bibitem{fulcher1997}
Fulcher, K.Y., White, P.D.: Randomised controlled trial of graded exercise in
  patients with the chronic fatigue syndrome. Bmj  314(7095),  1647--1652
  (1997)

\bibitem{giangiacomo2008}
Giangiacomo, M.C.P.B., Navas, A.L.G.P.: A influ{\^e}ncia da mem{\'o}ria
  operacional nas habilidades de compreens{\~a}o de leitura em escolares de
  4{\textordfeminine} s{\'e}rie influence of working memory in reading
  comprehension in 4th grade students. Sociedade Brasileira de Fonoaudiologia
  13(1),  69--74 (2008)

\bibitem{graesser2011}
Graesser, A.C., McNamara, D.S., Kulikowich, J.M.: Coh-metrix providing
  multilevel analyses of text characteristics. Educational Researcher  40(5),
  223--234 (2011)

\bibitem{graesser2004}
Graesser, A.C., McNamara, D.S., Louwerse, M.M., Cai, Z.: Coh-metrix: Analysis
  of text on cohesion and language. Behavior research methods, instruments, \&
  computers  36(2),  193--202 (2004)

\bibitem{hancke2012}
Hancke, J., Vajjala, S., Meurers, D.: Readability classification for german
  using lexical, syntactic, and morphological features. In: Proceedings of
  COLING. pp. 1063--1080 (2012)

\bibitem{hovy2010}
Hovy, E., Lavid, J.: Towards a ‘science’of corpus annotation: a new
  methodological challenge for corpus linguistics. International journal of
  translation  22(1),  13--36 (2010)

\bibitem{kato1985}
Kato, M.: O aprendizado da leitura. Martins Fontes (1985)

\bibitem{kato1986}
Kato, M.A.: No mundo da escrita: uma perspectiva psicoling{\"u}{\'\i}stica,
  vol.~9. Editora {\'A}tica (1986)

\bibitem{krieger2012}
Krieger, M.d.G.: Dicion{\'a}rios para o ensino de l{\'\i}ngua materna:
  princ{\'\i}pios e crit{\'e}rios de escolha. Revista L{\'\i}ngua\&Literatura
  7(10-11),  101--112 (2012)

\bibitem{landiskoch1977}
Landis, J.R., Koch, G.G.: The measurement of observer agreement for categorical
  data. Biometrics  33(1),  pp. 159--174 (1977)

\bibitem{lennon2004}
Lennon, C., Burdick, H.: The lexile framework as an approach for reading
  measurement and success. Electronic publication on www.lexile.com  (2004)

\bibitem{lopucki2014}
LoPucki, L.M.: System and method for enhancing comprehension and readability of
  legal text (2014), {US} Patent 8,794,972

\bibitem{maia2001}
Maia, M.: Gram{\'a}tica e parser. Boletim da ABRALIN  1(26) (2001)

\bibitem{maia2011}
Maia, M.: Efeitos do status argumental e de segmenta{\c{c}}{\~a}o no
  processamento de sintagmas preposicionais em portugu{\^e}s brasileiro.
  Cadernos de Estudos Ling{\"u}{\'\i}sticos  50(1) (2011)

\bibitem{maia2005}
Maia, M., FINGER, I.: Processamento da linguagem. Pelotas: Educat  (2005)

\bibitem{martins1996}
Martins, T.B., Ghiraldelo, C.M., Nunes, M.d.G.V., de~Oliveira~Junior, O.N.:
  Readability formulas applied to textbooks in brazilian portuguese. Icmsc-Usp
  (1996)

\bibitem{maziero2008aic}
Maziero, E.G., Pardo, T.A.S., Alu{\'\i}sio, S.M.: Ferramenta de an{\'a}lise
  autom{\'a}tica de inteligibilidade de c{\'o}rpus (aic). Tech. rep. (2008)

\bibitem{navas2009}
Navas, A.L.G.P., Pinto, J.C.B.R., Dellisa, P.R.R.: Avan{\c{c}}os no
  conhecimento do processamento da flu{\^e}ncia em leitura: da palavra ao texto
  improvements in the knowledge of the reading fluency processing: from word to
  text. Sociedade Brasileira de Fonoaudiologia  14(3),  553--9 (2009)

\bibitem{oreilly2004}
O'Reilly, T., Sinclair, G., McNamara, D.S.: istart: A web-based reading
  strategy intervention that improves students's science comprehension. In:
  CELDA. pp. 173--180 (2004)

\bibitem{petersen2009}
Petersen, S.E., Ostendorf, M.: A machine learning approach to reading level
  assessment. Computer speech \& language  23(1),  89--106 (2009)

\bibitem{desalles2006}
de~Salles, J.S.F., Parente, M.A.d.M.P.: Heterogeneidade nas estrat{\'e}gias de
  leitura/escrita em crian{\c{c}}as com dificuldades de leitura e escrita.
  Psico  37(1),  83--90

\bibitem{san2014}
San~Norberto, E.M., G{\'o}mez-Alonso, D., Trigueros, J.M., Quiroga, J., Gualis,
  J., Vaquero, C.: Readability of surgical informed consent in spain.
  Cirug{\'\i}a Espa{\~n}ola  92(3),  201--207 (2014)

\bibitem{scartonaluisio2010}
Scarton, C., Alu{\'i}sio, S.: {An{\'a}lise da Inteligibilidade de textos via
  ferramentas de Processamento de L{\'i}ngua Natural: adaptando as m{\'e}tricas
  do Coh-Metrix para o Portugu{\^e}s}. Linguam{\'a}tica  2(1),  45--62 (2010)

\bibitem{sheehan2013}
Sheehan, K.M., Flor, M., Napolitano, D.: A two-stage approach for generating
  unbiased estimates of text complexity. In: Proceedings of the Workshop on
  Natural Language Processing for Improving Textual Accessibility. pp. 49--58
  (2013)

\bibitem{stenner1996}
Stenner, A.J.: Measuring reading comprehension with the lexile framework
  (1996)

\bibitem{tong2002}
Tong, S., Koller, D.: Support vector machine active learning with applications
  to text classification. The Journal of Machine Learning Research  2,  45--66
  (2002)

\bibitem{vajjala2014}
Vajjala, S., Meurers, D.: Readability assessment for text simplification: From
  analysing documents to identifying sentential simplifications. International
  Journal of Applied Linguistics  165(2),  194--222 (2014)

\end{thebibliography}
